\documentclass[letterpaper, 10 pt, conference]{ieeeconf}  
\IEEEoverridecommandlockouts

\usepackage[ruled,vlined]{algorithm2e}

\usepackage{xspace}

\newcommand{\CertiFGO}{\textnormal{\texttt{Certi-FGO}}\xspace}
\newcommand{\CertiGNC}{\textnormal{\texttt{Certi-GNC}}\xspace}
\newcommand{\GNCLocal}{\textnormal{\texttt{GNC-Local}}\xspace}

\usepackage{cite}
\usepackage{bm}

\usepackage{comment}

\usepackage[toc,acronyms,nopostdot,nogroupskip, nonumberlist]{glossaries-extra} 
\MFUhyphentrue 

\glssetcategoryattribute{general}{glossdesc}{title}
\glssetcategoryattribute{abbreviation}{glossdesc}{title}
\makeatletter
\newglossarystyle{long-initcapsdesc}{%
  \setglossarystyle{long}%
    \renewcommand{\glossentry}[2]{%
    \glsentryitem{##1}\glstarget{##1}{\glossentryname{##1}} &
    \protected@edef\thisdesc{\glsentrydesc{##1}}%
    \xcapitalisewords{\thisdesc}\glspostdescription\space ##2\tabularnewline
  }%
}
\makeatother
\makeglossaries
\setglossarystyle{index}

\newabbreviation[shortplural={POPs}, longplural={polynomial optimization problems}]{POP}{POP}{polynomial optimization problem}
\newabbreviation[shortplural={SDPs}, longplural={semidefinite programs}]{SDP}{SDP}{semidefinite program}

\newabbreviation[shortplural={SOS}, longplural={sums of squares}]{SOS}{SOS}{sum of squares}

\newabbreviation{GNC}{GNC}{graduated non-convexity}

\newabbreviation[shortplural={QCQPs}, longplural={quadratically constrained quadratic programs}]{QCQP}{QCQP}{quadratically constrained quadratic program}

\newabbreviation{RMSE-ATE}{RMSE-ATE}{root-mean-square absolute trajectory error}

\newabbreviation{ADAPT}{ADAPT}{adaptive trimming}

\newabbreviation{PSD}{PSD}{positive semidefinite}

\newabbreviation{PnP}{PnP}{perspective-n-point}

\newabbreviation{CORA}{CORA}{certifiably correct range-aided SLAM}

\newabbreviation{NLS}{NLS}{nonlinear least-squares}

\newabbreviation{IRLS}{IRLS}{iteratively reweighted least squares}

\newabbreviation{BSS}{BSS}{blind source separation}

\newabbreviation{RL}{RL}{reinforcement learning}

\newabbreviation{RJDM}{RJDM}{radar jamming decision-making}

\newabbreviation{LORO}{LORO}{lobe-on-receive-only}

\newabbreviation{RPIP}{RPIP}{random pulse initial phases}

\newabbreviation{ATE}{ATE}{absolute trajectory error}

\newabbreviation{RPE}{RPE}{relative pose error}

\newabbreviation{TLS}{TLS}{total least squares}
\newabbreviation{GM}{GM}{Geman--McClure}

\newabbreviation{PGO}{PGO}{pose graph optimization}

\newabbreviation{BR}{BR}{Black-Rangarajan}

\newabbreviation{PPP}{PPP}{precise point positioning}

\newabbreviation{SLAM}{SLAM}{simultaneous localization and mapping}

\newabbreviation{SfM}{SfM}{structure from motion}

\newabbreviation{VIO}{VIO}{visual–inertial odometry}

\newabbreviation{ICP}{ICP}{iterative closest point}

\newabbreviation{FIM}{FIM}{Fisher information matrix}

\newabbreviation{CRB}{CRB}{Cramér-Rao bound}

\newabbreviation{CP}{CP}{correlated processing}

\newabbreviation{CPI}{CPI}{coherent processing interval}

\newabbreviation{IC}{IC}{interference cancellation}

\newabbreviation{MUSIC}{MUSIC}{multiple signal classification}

\newabbreviation{SAM}{SAM}{swept amplitude-modulation}

\newabbreviation{AM}{AM}{swept amplitude-modulation}

\newabbreviation{DBS}{DBS}{Doppler beam sharpening}

\newabbreviation{MTT}{MTT}{multiple target tracking}

\newabbreviation{LFM}{LFM}{linear frequency-modulated}

\newabbreviation{LOS}{LOS}{line-of-sight}

\newabbreviation[longplural={directions of arrival}]{DOA}{DOA}{direction of arrival}

\newabbreviation{SPJ}{SPJ}{self-protection jamming}

\newabbreviation{SOJ}{SOJ}{stand-off jamming}

\newabbreviation{FGO}{FGO}{factor graph optimization}

\newabbreviation{MLE}{MLE}{maximum likelihood estimation}

\newabbreviation[longplural={neural networks}]{NN}{NN}{neural network}

\newabbreviation{EJ}{EJ}{escort jamming}

\newabbreviation{JRC}{JRC}{joint radar-communication}

\newabbreviation{WLS}{WLS}{weighted least squares}

\newabbreviation{MAP}{MAP}{maximum a posteriori}

\newabbreviation{LM}{LM}{Levenberg-Marquardt}

\newabbreviation{ERP}{ERP}{effective radiated power}

\newabbreviation{LTI}{LTI}{linear time-invariant}

\newabbreviation[shortplural={KFs}, longplural={Kalman filters}]{KF}{KF}{Kalman filter}

\newabbreviation[shortplural={EKFs}, longplural={extended Kalman filters}]{EKF}{EKF}{extended Kalman filter}

\newabbreviation[shortplural={PFs}, longplural={particle filters}]{PF}{PF}{particle filter}

\newabbreviation[shortplural={UKFs}, longplural={unscented Kalman filters}]{UKF}{UKF}{unscented Kalman filter}

\newabbreviation[shortplural={GSFs}, longplural={Gaussian sum filters}]{GSF}{GSF}{Gaussian sum filter}

\newabbreviation[shortplural={DOFs}, longplural={degrees of freedom}]{DOF}{DOF}{degree of freedom}

\newabbreviation[shortplural={MMFTs}, longplural={micro-motion false targets}]{MMFT}{MMFT}{micro-motion false target}

\newabbreviation[shortplural={TFTs}, longplural={translational false targets}]{TFT}{TFT}{translational false target}

\newabbreviation[shortplural={VAEs}, longplural={variational autoencoders}]{VAE}{VAE}{variational autoencoder}

\newabbreviation[shortplural={PRIs}, longplural={pulse repetition intervals}]{PRI}{PRI}{pulse repetition interval}

\newabbreviation[shortplural={PTs}, longplural={physical targets}]{PT}{PT}{physical target}

\newabbreviation{FFT}{FFT}{fast Fourier transform}

\newabbreviation[shortplural={MMs}, longplural={max mixtures}]{MM}{MM}{max mixtures}

\newabbreviation{MMSE}{MMSE}{minimum mean-square error}

\newabbreviation{DCS}{DCS}{dynamic covariance scaling}

\newabbreviation{SC}{SC}{switchable constraints}

\newabbreviation{DA}{DA}{data association}

\newabbreviation{JSR}{JSR}{jamming-to-signal ratio}

\newabbreviation{JNR}{JNR}{jamming-to-noise ratio}

\newabbreviation{RCS}{RCS}{radar cross-section}

\newabbreviation[longplural={state-space models}]{SSM}{SSM}{state-space model}

\newabbreviation{CDF}{CDF}{cumulative distribution function}

\newabbreviation{CS}{CS}{compressed sensing}

\newabbreviation{VI}{VI}{variational inference}

\newabbreviation{MSE}{MSE}{mean squared error}

\newabbreviation{AI}{AI}{artificial intelligence}

\newabbreviation{MCRB}{MCRB}{misspecified CRB}

\newabbreviation{EW}{EW}{electronic warfare}

\newabbreviation{CW}{CW}{continuous-wave}

\newabbreviation{IF}{IF}{influence function}

\newabbreviation{HS}{HS}{homogeneity and separation scores}

\newabbreviation[shortplural={TOIs}, longplural={targets of interest}]{TOI}{TOI}{target of interest}

\newabbreviation[shortplural={ECMs}, longplural={electronic countermeasures}]{ECM}{ECM}{electronic countermeasure}

\newabbreviation[shortplural={ECCMs}, longplural={electronic counter-countermeasures}]{ECCM}{ECCM}{electronic counter-countermeasure}

\newabbreviation[shortplural={FTs}, longplural={false targets}]{FT}{FT}{false target}

\newabbreviation[shortplural={PDs}, longplural={pulse Dopplers}]{PD}{PD}{pulse Doppler}

\newabbreviation[shortplural={FTGs}, longplural={false target generators}]{FTG}{FTG}{false target generator}

\newabbreviation[shortplural={FDAs}, longplural={frequency diverse arrays}]{FDA}{FDA}{frequency diverse array}

\newabbreviation[shortplural={SARs}, longplural={synthetic aperture radars}]{SAR}{SAR}{synthetic aperture radar}

\newabbreviation[shortplural={DRFMs}, longplural={digital radio-frequency memories}]{DRFM}{DRFM}{digital radio frequency memory}

\newabbreviation[shortplural={RGPOs}, longplural={range gate pull-offs}]{RGPO}{RGPO}{range gate pull-off}

\newabbreviation[shortplural={RGPIs}, longplural={range gate pull-ins}]{RGPI}{RGPI}{range gate pull-in}

\newabbreviation[shortplural={VGPOs}, longplural={velocity gate pull-offs}]{VGPO}{VGPO}{velocity gate pull-off}

\newabbreviation[shortplural={VGPIs}, longplural={velocity gate pull-ins}]{VGPI}{VGPI}{velocity gate pull-in}

\newabbreviation[shortplural={RVGPOs}, longplural={range-velocity gate pull-offs}]{RVGPO}{RVGPO}{range-velocity gate pull-off}

\newabbreviation[shortplural={RVGPIs}, longplural={range-velocity gate pull-ins}]{RVGPI}{RVGPI}{range-velocity gate pull-in}

\newabbreviation[shortplural={ISRJs}, longplural={interrupted-sampling repeater jammings}]{ISRJ}{ISRJ}{interrupted-sampling repeater jamming}

\newabbreviation{CRDJ}{CRDJ}{crosspulse repeater deception jamming}

\newabbreviation[shortplural={MHTs}, longplural={multiple hypothesis trackings}]{MHT}{MHT}{multiple hypothesis tracking}

\newabbreviation[shortplural={RNNs}, longplural={recurrent neural networks}]{RNN}{RNN}{recurrent neural network}

\newabbreviation[shortplural={SJNRs}, longplural={signal-to-jammer noise ratios}]{SJNR}{SJNR}{signal-to-jammer noise ratio}

\newabbreviation[shortplural={CNNs}, longplural={convolutional neural networks}]{CNN}{CNN}{convolutional neural network}

\newabbreviation[shortplural={LSTMs}, longplural={long short-term memories}]{LSTM}{LSTM}{long short-term memory}

\newabbreviation[shortplural={TDOAs}, longplural={time differences of arrival}]{TDOA}{TDOA}{time difference of arrival}

\newabbreviation[shortplural={RFSs}, longplural={random finite sets}]{RFS}{RFS}{random finite set}

\newabbreviation[shortplural={SNRs}, longplural={signal-to-noise ratios}]{SNR}{SNR}{signal-to-noise ratio}

\newabbreviation[shortplural={SJRs}, longplural={signal-to-jammer ratios}]{SJR}{SJR}{signal-to-jammer ratio}

\newabbreviation[shortplural={OFDMs}, longplural={orthogonal frequency-division multiplexings}]{OFDM}{OFDM}{orthogonal frequency-division multiplexing}

\newabbreviation[shortplural={PRFs}, longplural={pulse repetition frequencies}]{PRF}{PRF}{pulse repetition frequency}

\newabbreviation[shortplural={SIMOs}, longplural={single-input multiple-outputs}]{SIMO}{SIMO}{single-input multiple-output}

\newabbreviation[shortplural={MIMOs}, longplural={multiple-input multiple-outputs}]{MIMO}{MIMO}{multiple-input multiple-output}

\newabbreviation[shortplural={GLRTs}, longplural={generalized likelihood ratio tests}]{GLRT}{GLRT}{generalized likelihood ratio test}

\newabbreviation[shortplural={AGCs}, longplural={automatic gain controls}]{AGC}{AGC}{automatic gain control}

\newabbreviation[shortplural={MLs}, longplural={machine learnings}]{ML}{ML}{machine learning}

\newabbreviation[shortplural={UAVs}, longplural={unmanned aerial vehicles}]{UAV}{UAV}{unmanned aerial vehicle}

\newabbreviation[shortplural={GNSSs}, longplural={global navigation satellite systems}]{GNSS}{GNSS}{global navigation satellite system}

\newabbreviation[shortplural={LICQs}, longplural={linear independence constraint qualifications}]{LICQ}{LICQ}{linear independence constraint qualifications}

\newabbreviation[shortplural={KKTs}, longplural={Karush-Kuhn-Tucker}]{KKT}{KKT}{Karush-Kuhn-Tucker}

\newabbreviation[shortplural={BMs}, longplural={Burer-Monteiros}]{BM}{BM}{Burer-Monteiro}

\usepackage[dvipsnames]{xcolor} 

\definecolor{myComment}{rgb}{0.1, 0.1, 1.0} 

\SetKwComment{Comment}{\textcolor{myComment}{//\ }}{} 
\SetCommentSty{mycommentstyle}    

\usepackage{graphics} 
\usepackage{epsfig} 
\usepackage{amsmath} 
\usepackage{amssymb}  
\usepackage{amsmath,amssymb,amsfonts}
\usepackage{graphicx}
\usepackage{textcomp}

\usepackage{amsfonts}  
\usepackage{mathrsfs}  
\usepackage{float}
\usepackage{lettrine}

\usepackage[utf8]{inputenc}
\usepackage{booktabs}    
\usepackage{caption}     
\usepackage{graphicx} 

\newtheorem{remark}{Remark}
\usepackage{dblfloatfix}  
\usepackage{placeins}     
\usepackage{wrapfig,lipsum,booktabs}
\usepackage{float}
\usepackage{placeins}  
\usepackage{balance}
\usepackage{afterpage}

\usepackage{subcaption}
\usepackage{tabularx} 

 \renewcommand{\boldsymbol}{\mathbf}

\newcommand{\cp}[1]{\ifmmode {\mathcal{#1}}\else ${\mathcal{#1}}$\fi}
\usepackage{array}
\newcolumntype{P}[1]{>{\centering\arraybackslash}p{#1}}

\newcommand{\bA}{\boldsymbol{A}}

\newcommand{\bQ}{\boldsymbol{Q}}

\newcommand{\bS}{\boldsymbol{S}}

\newcommand{\bX}{\boldsymbol{X}}
\newcommand{\bY}{\boldsymbol{Y}}
\newcommand{\bZ}{\boldsymbol{Z}}

\newcommand{\bb}{\boldsymbol{b}}

\newcommand{\blambda}{\bm{\lambda}}

\usepackage{nccmath}

\def\credrev2{\textcolor{red}}
\def\credrev{\textcolor{red}}

\definecolor{darkgreen}{rgb}{0., 0.4, 0.}
\definecolor{amber}{rgb}{1.0, 0.49, 0.0}
\definecolor{orange}{rgb}{1.0, 0.4, 0.0}

\usepackage{hyperref}
\hypersetup{colorlinks=true,allcolors=blue}
\makeatletter
\let\NAT@parse\undefined
\makeatother
\begin{document}

\title{\LARGE \bf Implementing Robust M-Estimators with Certifiable Factor Graph Optimization}

\author{%
Zhexin Xu, Hanna Jiamei Zhang, Helena Calatrava, Pau Closas, David M. Rosen
\thanks{The authors are with Institute for Experiential Robotics, Northeastern University, 360 Huntington Ave, Boston, MA 02115, USA. \texttt{\{xu.zhex, zhang.hanna, calatrava.h, closas, d.rosen\}@northeastern.edu}.  We acknowledge the support of the Natural Sciences and Engineering Research Council of Canada (NSERC). Nous remercions le Conseil de recherches en sciences naturelles et en génie du Canada (CRSNG) de son soutien. This work has been partially supported by the National Science Foundation under Awards 1845833, 2326559 and 2530870, and by MIT Lincoln Laboratory through Air Force Contract FA8702-15-D-0001.}
}

\maketitle

\begin{abstract}

Parameter estimation in robotics and computer vision faces formidable challenges from both outlier contamination and nonconvex optimization landscapes. While M-estimation addresses the problem of outliers through robust loss functions, it creates severely nonconvex problems that are difficult to solve globally. \emph{Adaptive reweighting} schemes provide one particularly appealing strategy for implementing M-estimation in practice: these methods solve a sequence of simpler weighted least squares (WLS) subproblems, enabling both the use of standard least squares solvers and the recovery of higher-quality estimates than simple local search. However, adaptive reweighting still crucially relies upon solving the inner WLS problems effectively, a task that remains challenging in many robotics applications due to the intrinsic nonconvexity of many common parameter spaces (e.g.\ rotations and poses).

In this paper, we show how one can easily implement adaptively-reweighted M-estimators with \emph{certifiably correct} solvers for the inner WLS subproblems using only fast \emph{local} optimization over smooth manifolds. 
Our approach exploits recent work on \emph{certifiable factor graph optimization} to provide global optimality certificates for the inner WLS subproblems while seamlessly integrating into existing factor graph-based software libraries and workflows.  Experimental evaluation on  pose-graph optimization and landmark SLAM tasks demonstrates that our adaptively reweighted certifiable estimation approach provides higher-quality estimates than alternative local search-based methods, while scaling tractably to realistic problem sizes.

\end{abstract}

\section{Introduction}

Current state-of-the-art approaches to parameter estimation in robotics and computer vision typically formalize and solve this task using \gls{MLE} \cite{dellaert2017factor,Barfoot_2017}.  This approach is appealing for both its conceptual simplicity, and for the strong statistical performance guarantees that \gls{MLE} affords  \cite{Cover2006Elements}.  However, \gls{MLE} faces two critical challenges.  First, it is  typically implemented using \emph{local} optimization methods that are highly sensitive to initialization when applied to nonconvex problems \cite{nocedal1999numerical}.  Second, basic \gls{MLE} estimators have \emph{zero breakdown point}, meaning that even a vanishingly small fraction of outlier contamination in the data can produce arbitrarily poor \gls{MLE} estimates \cite{ronchetti2009robust}.

\emph{M-estimation} \cite{ronchetti2009robust} addresses the outlier problem by replacing the standard negative log-likelihood loss used in basic \gls{MLE} with \emph{robust} alternatives that are designed to attenuate the ill effects of outlier measurements on the resulting estimates. While this approach can dramatically improve robustness to outliers, achieving this effective attenuation typically requires the use of \emph{nonconvex} loss functions that substantially exacerbate the nonconvexity (and hence sensitivity to initialization) already present in \gls{MLE}.  

Implementing M-estimation in practice thus frequently entails a tradeoff between performance guarantees and computational tractability.  On one hand, global optimization methods (such as branch-and-bound or global polynomial optimization techniques) \cite{lasserre2001global,yang2020one} can guarantee the recovery of correct (i.e.\ \emph{globally optimal}) M-estimates, but are typically intractably expensive to apply to large-scale problems.  Conversely, \emph{local} optimization remains computationally efficient, but determining a suitable high-quality initialization becomes even more challenging in the presence of potential outlier contamination \cite{carlone2015initialization}.

\begin{figure}[t]
\centering
\includegraphics[width=\columnwidth]{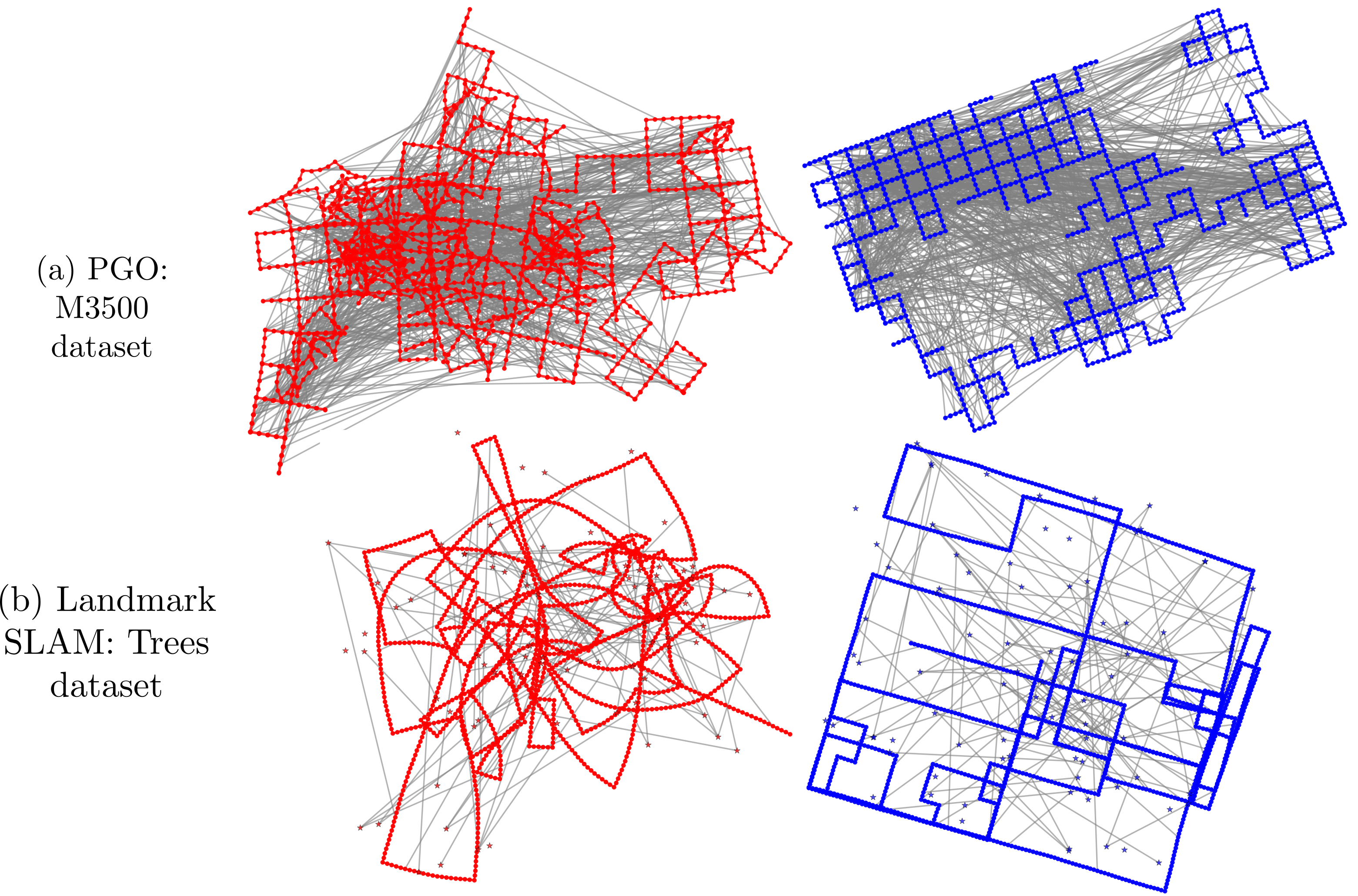}  
\caption{Examples of solutions obtained with \GNCLocal (left) and the proposed \CertiGNC framework (right), 
in the presence of 30\,\% outlier loop closures (shown in grey) for a) PGO and b) landmark SLAM.
}
\label{fig:intro_SLAM}
\vspace{-0.5cm}
\end{figure}

\emph{Adaptive reweighting} schemes provide one particularly appealing strategy for implementing M-estimation.  These methods proceed by solving a \emph{sequence} of \gls{WLS} problems, which is advantageous because the latter are frequently easier to optimize than robust formulations (and in fact often take the form of a standard \gls{MLE} for which a fast solver is already available).  Adaptive reweighting thus provides a computationally efficient approach to implementing M-estimation using standard solvers, while also typically recovering substantially better estimates than direct application of simple local search. However, this strategy crucially relies upon solving the \emph{inner} \gls{WLS}, a task that is itself still challenging in many robotics applications, due to the inherent nonconvexity of the underlying parameter spaces (e.g.\ rotation and pose manifolds) \cite{SLAMHandbookTheoryChp}.  While some recent work has demonstrated the use of convex relaxation-based \emph{certifiably correct} methods to perform this inner \gls{WLS} optimization \cite{yang2020graduated}, at present certifiable methods are typically hand-designed for specific problem classes (e.g.\ rotation averaging or \gls{PGO}), and thus this strategy -- while highly effective -- is limited to those specific applications for which an existing certifiable optimizer is already available \cite{SLAMHandbookTheoryChp}.

In this paper, we show how to easily implement a broad class of adaptively reweighted M-estimators with certifiably correct inner \gls{WLS} solvers using only fast local optimization over smooth manifolds, building on the certifiable factor graph framework of \cite{certi_fgo}.
Our main contributions are:
\begin{enumerate}

\item We propose a robust implementation\footnote{Code available at: \url{https://github.com/NEU-RAL/Certi-GNC}} of certifiable estimation by combining an adaptive reweighting framework with certifiable factor graphs \cite{certi_fgo}, without requiring hand-designed, problem-specific certifiable solvers.

\item We show on pose-graph and landmark SLAM that our method consistently yields higher-quality solutions than alternative M-estimation methods, while remaining computationally tractable on realistic problems.

\end{enumerate}

\section{Review of Robust Estimation}\label{sec:robust_est_theory}

This section reviews fundamentals of robust estimation, focusing on the use of adaptive reweighting schemes to solve the M-estimation problem~\cite{SLAMHandbookOutliersChp}.

M-estimation computes a state estimate $\bm{x}^\ast$ according to:
\begin{align}\label{eq:map2mest}
    \bm{x}^\ast = \arg \min_{\bm{x}} \sum_i \rho(r_i(\bm{x})),
\end{align}
%
%
where $\rho(\cdot)$ is a \emph{robust} loss function that grows sub-quadratically for large errors.  Ensuring hard rejection of very large residuals requires a loss whose gradient \emph{vanishes} for large errors, which is necessarily nonconvex; this additional nonconvexity makes robust M-estimation harder than its non-robust counterpart.

One can address this challenge by directly applying \emph{global} optimization to solve~\eqref{eq:map2mest}; however, this problem is NP-hard in general, and thus global methods quickly become intractable for large-scale SLAM or computer vision problems~\cite{SLAMHandbookOutliersChp,SLAMHandbookTheoryChp}. Consequently, practitioners often rely on local nonlinear methods, which are initialization-sensitive and prone to poor local minima under outlier contamination~\cite{carlone2015initialization}.

\SetKwFunction{Continue}{Continue}
\SetKwFunction{UpdateSchedule}{UpdateSchedule}
\SetKwFunction{ConvergenceWLS}{ConvergenceWLS}
\SetKwFunction{ConvergenceGNC}{ConvergenceGNC}
\begin{algorithm}[t]
\DontPrintSemicolon
\SetAlgoLined
\LinesNumbered
\caption{GNC for Robust M-estimation \cite{yang2020graduated}}
%
%
\label{alg:algorithm_gnc}
\KwIn{Initial estimate $\bm{x}^{(0)}$, initial weights $\bm{w}^{(0)}$, surrogate loss family $\rho_\mu(\cdot)$ 
}
\KwOut{Final estimate $\hat{\bm{x}}$}

\textbf{Initialize:} $\bm{x} \gets \bm{x}^{(0)}$, $\bm{w} \gets \bm{w}^{(0)}$, $\mu \gets \mu_0$\;

\Repeat{\textsc{ConvergenceGNC()}{} (Sec.~\ref{sec:iii:implementation_details})}{
  \Repeat{\textsc{ConvergenceWLS()}{} (Sec.~\ref{sec:iii:implementation_details})}{
    $\bm{x} \gets \displaystyle\arg\min_{\bm{x}} \sum_{i=1}^{n} w_i \, r_i^2(\bm{x})$\label{WLS_in_GNC}
    
      \For{$i \in [n]$}{
      \vspace{-.4cm}
  \begin{equation*}
  w_i \gets \arg\min_{w_i \in [0,1]} \,\Phi_{\rho_\mu}(w_i) + w_i r_i^2(\bm{x})\end{equation*} 
  \vspace{-0.5cm}
}
  }
  $\mu \gets$ \textsc{UpdateSchedule}{($\mu$)}
}
\Return $\hat{\bm{x}} \gets \bm{x}$\;
\end{algorithm}
\subsection{Reformulating M-estimation as Outlier Processes}\label{sec:br}

Rather than relying solely on global or purely local optimization to directly solve the robust problem~\eqref{eq:map2mest}, a practical alternative is to \emph{robustify} an existing non-robust estimator by solving a sequence of simpler \gls{WLS} subproblems via adaptive reweighting schemes.
The \gls{BR} duality~\cite{black1996unification} formalizes this strategy by describing an equivalence between solving the M-estimation problem by applying \textit{(i)} a robust loss to the measurement residuals and \textit{(ii)} a weighted, non-robust joint formulation over \((\bm{x}, \bm{w})\), where $\bm{w}$ are a set of auxiliary weights multiplying the quadratic residual terms.
This equivalence is presented in the following theorem:

\newtheorem{thm}{Theorem}
\begin{thm}\label{Theorem_1}
    (Black--Rangarajan Duality \cite{black1996unification}) 
    \textit{Given a robust loss $\rho(\cdot)$, define $\phi(z) := \rho(\sqrt{z})$. 
    If $\phi(z)$ satisfies $\lim_{z \to 0} \phi'(z) = 1$, 
    $\lim_{z \to \infty} \phi'(z) = 0$, and $\phi''(z) < 0$, 
    then the M-estimation problem in~\eqref{eq:map2mest} is equivalent to}
    \begin{equation} \label{eq:br_duality}
        \min_{\substack{\bm{x} \\ \bm{w}\in[0,1]^n}} 
        \sum_{i=1}^{n} \left[ w_i r_i^2(\bm{x}) + \Phi_\rho(w_i) \right],
    \end{equation}
    \textit{where \(w_i \in [0,1]\)
 are  auxiliary weights, and $\Phi_\rho(w_i)$ is an outlier process induced by $\rho(\cdot)$.}
\end{thm}

%
The conditions on $\rho(\cdot)$ are satisfied by most common robust loss functions~\cite{black1996unification}. 
One nice feature of this approach is that the optimal auxiliary weights can be interpreted as \emph{soft inlier indicators}. Note that this probabilistic interpretation is a byproduct of the \gls{BR} reformulation and is not available when minimizing a robust loss alone.

\subsection{IRLS via Alternating Minimization}\label{sec:2c}

Directly minimizing \emph{jointly} over $(\bm{x},\bm{w})$ in~\eqref{eq:br_duality} is frequently difficult. However, two \emph{partial} minimizations are often much more straightforward: \textit{(i)} for fixed $\bm{w}$, minimizing over $\bm{x}$ reduces to an ordinary \gls{WLS} problem; and \textit{(ii)} for fixed $\bm{x}$, the minimization over the weights decouples into $n$ independent one-dimensional problems on $[0,1]$, which for many losses admit closed-form solutions. Note that while this \gls{WLS} problem is easier to implement with standard NLS methods (Gauss--Newton, \gls{LM}) than the original problem, it remains challenging for most robotics applications. Applying alternating minimization to the BR formulation (Theorem \ref{Theorem_1}) simply alternates these two steps and yields the classical \gls{IRLS} algorithm.
Because each partial minimization cannot increase the objective, \gls{IRLS} produces a monotonically nonincreasing cost sequence.
Nevertheless, the nonconvexity of robust redescending losses renders the convergence of \gls{IRLS} highly sensitive to initialization~\cite{SLAMHandbookOutliersChp}.

\subsection{GNC as a Continuation Strategy}\label{sec:2d}

\begin{figure}[t]
\centering
        \includegraphics[width=\columnwidth]{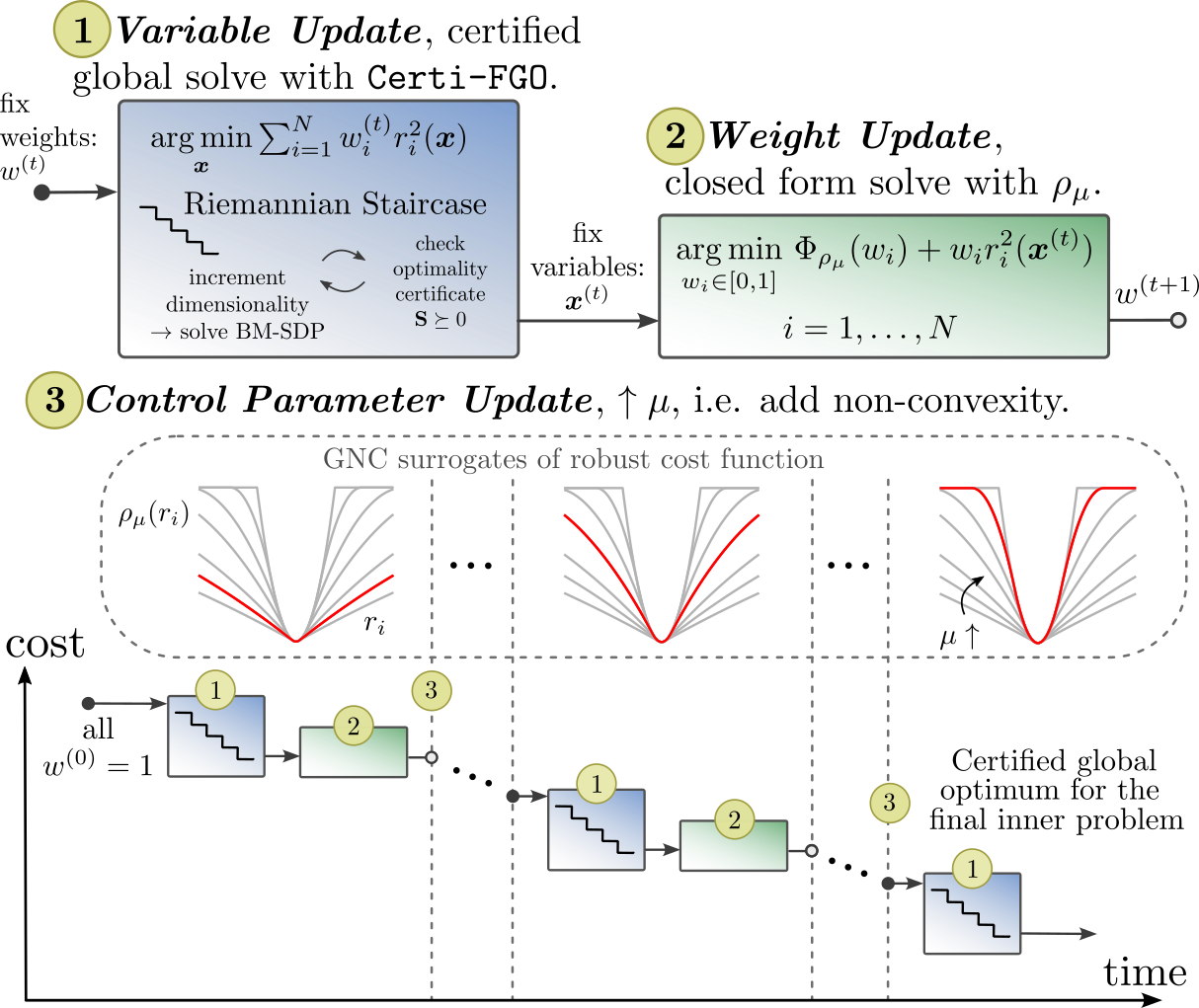}
        \caption{Overview of our \CertiGNC framework: \CertiFGO~\cite{certi_fgo} provides 1) certifiable global inner solves (i.e. non-linear weighted least squares (WLS) problem) within a graduated non-convexity (GNC) framework. The control parameter~$\mu$ defines convex surrogates $\rho_\mu(r_i)$ of the target non-convex robust loss, yielding a sequence of WLS problems with 2) weights $w_i\in[0,1]$ updated via. closed form solve from residuals $r_i$. 3) A truncated least squares (TLS) robust loss function is used here: $\mu$ is increased to gradually recover non-convexity, with $\mu\!\to\!\infty$ recovering original truncation. 
        }
        \label{fig:intro}
        \vspace{-0.5cm}
\end{figure}

\Gls{GNC} is a continuation scheme that aims to reduce sensitivity to initialization by introducing a homotopy from an ``easy" convex surrogate to the ``hard" target robust loss, with each stage solved by 
\gls{WLS}. Compared with \gls{IRLS}~\cite{blake1987visual},
\gls{GNC} introduces a control parameter~$\mu$ that defines surrogate losses $\rho_\mu(\cdot)$, which are initially convex and gradually recover the original robust loss as $\mu$ is varied. This yields a sequence of \gls{WLS} subproblems with progressively increasing robustness (see Fig.~\ref{fig:intro}). The \gls{BR} duality extends naturally to $\rho_\mu(\cdot)$, 
resulting in the corresponding outlier processes $\Phi_{\rho_\mu}(w_i)$ (see for instance~\cite{SLAMHandbookOutliersChp} for some common explicit forms).
This leads to the three steps shown in Algorithm~\ref{alg:algorithm_gnc}.

\section{A Robust Implementation of Certifiable Factor Graph Optimization}

In the previous section, we showed that the robust estimation problem can be reduced to a sequence of non-robust problems. However, obtaining \emph{global} solutions to the resulting non-robust subproblems remains challenging for many tasks of interest in robotics and computer vision.

The central question of this work is: within robust adaptive weighting schemes, \textit{how can we solve the inner \gls{WLS} problem (see Algorithm~\ref{alg:algorithm_gnc}) so that it is both certifiably correct and broadly applicable?} 
We address this by solving this inner problem through \CertiFGO~\cite{certi_fgo}, a framework that makes certifiable estimation accessible in factor graph libraries (e.g., GTSAM~\cite{gtsam}) by exploiting the structural correspondence between factor graphs and block-separable \glspl{QCQP}.

In the remainder of this section, we introduce Shor’s convex relaxation of these block-separable \glspl{QCQP}, yielding a \gls{SDP} that serves as our certifiable surrogate, and show how to implement and solve the resulting relaxations efficiently by only using standard local \gls{FGO} routines. We therefore preserve the practicality of \gls{FGO} and equip it with certifiable guarantees.

\subsection{Shor's Relaxation of QCQPs} \label{subsec:shors}

A growing body of recent work has shown that hard nonconvex \gls{MLE} problems can be solved by expressing them as a \gls{QCQP}, relaxing them to \gls{SDP}, and then applying \gls{BM} factorization \cite{burer2003nonlinear} to handle large problem instances efficiently~\cite{SLAMHandbookTheoryChp,papalia2024overview}.

Let us assume that we have an estimation problem in the form of a  \gls{QCQP}:
\begin{equation}\label{eq:qcqp}f^\ast_{\text{QCQP}}=
\min_{\bX\in\mathbb{R}^{n\times d}}\; \langle \bQ,\bX\bX^\top\rangle
\quad\text{s.t.} \langle \bA_i,\bX\bX^\top\rangle=b_i,\;\forall i\in[m],
\end{equation}
where $\bQ\in\mathbb{S}^n$, $\bA_i\in\mathbb{S}^{n}$ for all $i\in[m]$, $\bb\in\mathbb{R}^m$, and $\mathbb{S}^n$ denotes the set of $n\times n$ symmetric matrices.

To address NP-hardness of~\eqref{eq:qcqp}, Shor’s relaxation \cite{shor1987quadratic} replaces \(\bX\bX^\top\) with a generic \gls{PSD} matrix $\bZ\in\mathbb{S}^{n}_+$, removing the implicit rank-$d$ constraint to yield the convex \gls{SDP} relaxation:

\begin{equation}\label{eq:sdp}
f^\ast_{\text{SDP}} =\min_{\bZ\in\mathbb{S}^{n}_+}\; \langle \bQ,\bZ\rangle
\quad\text{s.t.} \langle \bA_i,\bZ\rangle=b_i,\;\forall i\in[m].
\end{equation}
Note that \eqref{eq:sdp} is convex, and can thus be solved to global optimality using e.g.\ interior-point methods. Furthermore, its optimal value always \emph{lower bounds} the optimal value of~\eqref{eq:qcqp}:
\(f^\ast_{\text{SDP}}\leq f^\ast_{\text{QCQP}}\).
Finally, note that if a minimizer \(\mathbf{Z}^\ast\) of~\eqref{eq:sdp} happens to have a rank-\(d\) factorization of the form \(\mathbf{Z}^\ast=\mathbf{X}^\ast{\mathbf{X}^\ast}^{\!\top}\),
then \(\mathbf{X}^\ast\) is in fact a \emph{global} minimizer of~\eqref{eq:qcqp}.

\begin{remark}\label{rem:solution-recovery}
Crucially, a large body of recent work has shown that this exact solution recovery often occurs when the measurement noise is small~\cite{rosen2019se, SLAMHandbookTheoryChp}.
\end{remark}

\subsection{Efficiently Solving the \gls{SDP} Relaxation via Low-Rank Factorization and Riemannian Staircase} \label{subsec:low_rank_staircase}
Rather than solving \eqref{eq:sdp} directly, we can exploit its low-rank structure to reduce computational complexity for large-scale problems.
We assume a factorization of the form \(\bZ=\bY\bY^\top\) for some \(\bY\in\mathbb{R}^{n\times p}\) and \( d < p\ll n\), yielding the \textit{lifted} \emph{rank-\(p\)} \gls{BM} factorization~\cite{burer2003nonlinear}:
\begin{equation}\label{eq:burer_monteiro}
\min_{\bY\in\mathbb{R}^{n\times p}}\; \langle \bQ,\bY\bY^\top\rangle
\quad\text{s.t.}\quad \langle \bA_i,\bY\bY^\top\rangle=b_i,\;\forall i\in[m],
\end{equation}
which automatically enforces positive semidefiniteness and reduces the decision space from roughly \(n^2\) to \(np\) variables (where $p \ll n $), at the cost of reintroducing nonconvexity.
Note that setting $p=d$ recovers the original QCQP~\eqref{eq:qcqp}. 
Moreover, a candidate $\bZ = \bY\bY^\top$ for a solution of \eqref{eq:sdp} construted from a \emph{local} minimizer $\bY$ of \eqref{eq:burer_monteiro} can still be checked for \emph{global} optimality using \eqref{eq:sdp}'s \gls{KKT} conditions.
\begin{remark}\label{rem:low-rank-justification}
This low-rank factorization approach is justified by the fact that \eqref{eq:sdp} is known to admit low-rank solutions for many problems of interest under mild conditions~\cite{rosen2019se,SLAMHandbookTheoryChp}.
\end{remark}

Through comparison of the \gls{KKT} conditions of the \gls{BM} and \gls{SDP} problems, one can show that a \gls{BM} solution can be certified as globally optimal by checking the positive semidefiniteness of the \textit{certificate matrix} \cite{rosen2020Scalable}:

\begin{equation}\label{equation: definition_of_certificate}
\bS \triangleq  \bQ + \mathcal{A}^\ast(\blambda) = \bQ + \sum_{i=1}^{m} \lambda_i \bA_i,
\end{equation}
where $\mathcal{A}^*:\mathbb{R}^m \to \mathbb{S}^n$ denotes the adjoint operator.
For a \gls{KKT} point $\bY$ of the rank-$p$ \gls{BM} factorization with multipliers $\blambda$, the matrix $\bS$ in~\eqref{equation: definition_of_certificate} serves as a certificate of optimality: if $\bS\succeq 0$, then $\bZ=\bY\bY^\top$ is globally optimal for the \gls{SDP} (and, when $p=d$, also solves the QCQP); otherwise, the eigenvector associated with the smallest eigenvalue of $\bS$ gives a direction of negative curvature. The \emph{Riemannian Staircase}~\cite{boumal2016non,Boumal2015RS} wraps this certify-or-lift step by solving a sequence of low-rank problems: parameterize $\bZ=\bY\bY^\top$ at rank $p$, solve the rank-$p$ manifold subproblem, recover multipliers to form $\bS$, and either certify or increase the rank to $p{+}1$ by augmenting $\bY$ along the minimum-eigenvector direction. In practice, one or two staircase steps typically suffice~\cite{papalia2024overview}.

\subsection{Certifiable Estimation for Factor Graphs}\label{subsec:certi-fgo}
We begin by considering the following general \gls{MLE} problem: jointly minimizing a sum of \textit{data fitting} terms with \textit{sparse} dependencies over a collection of variables,
\begin{equation}\label{eq:fgo_opt_prob}
\begin{aligned}
&\min_{\bm{x}_i \in \mathcal{M}_i} 
&&\sum_{k=1}^u l_k(\bm{x}_{S_k}), 
\end{aligned}
\end{equation}
where $\mathcal{M}_i$ is the domain of $\bm{x}_i$ and each summand $l_k$ depends \textit{only} upon the subset of variables $\bm{x}_{S_k}$ indexed by $S_k\subseteq [n]$. 

\begin{figure}[t]
\centering
        \includegraphics[width=1\columnwidth]{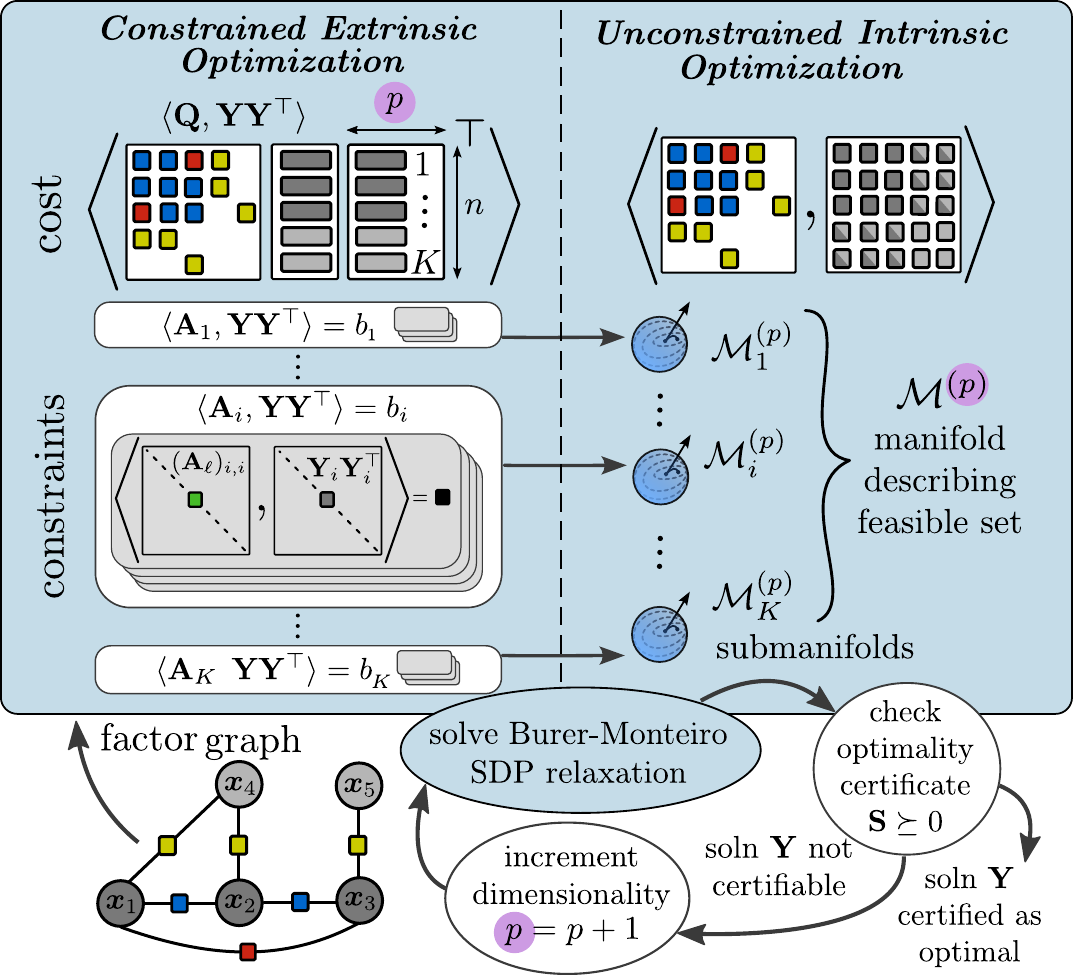}
        \caption{Illustration of the \CertiFGO framework showing the transformation from constrained to unconstrained optimization leveraging underlying sparse factor graph structure. The \gls{BM} variable $\mathbf{Y}$ is partitioned into $K$ block rows $\mathbf{Y}_i$ corresponding to variables $\bm{x}_i$ in the factor graph. The preserved block structure is apparent in this \gls{BM} form of the factor graph \gls{MLE}~\eqref{equation:bm_fgo}, in that each constraint block $(\mathbf{A}_{\ell})_{i,i}$ acts locally on a single block $\mathbf{Y}_i$, defining individual submanifolds $\mathcal{M}_i^{(p)}$. Sparse separable structure is also present in the objective, though not explicitly visualized. The constrained optimization problem (Eq.~\eqref{equation:bm_fgo}, left) is reformulated as unconstrained Riemannian optimization over the product manifold $\mathcal{M}^{(p)} = \mathcal{M}_1^{(p)} \times \cdots \times \mathcal{M}_K^{(p)}$ (Eq.~\eqref{equation: intrinsic_optimization_formulation}, right), enabling efficient solution using standard manifold optimization techniques.} \label{fig:certi-fgo}
        \vspace{-0.5cm}
\end{figure}

This sparse dependency structure of~\eqref{eq:fgo_opt_prob} admits a natural graphical representation using \textit{factor graphs} \cite{dellaert2017factor}. Formally the factor graph associated with \eqref{eq:fgo_opt_prob} is the bipartite graph $\mathcal{G} = (\mathcal{V}, \mathcal{F}, \mathcal{E})$ in which: 
\begin{enumerate}
\item\textit{variable nodes} $\mathcal{V}\triangleq\{\bm{x}_1, \dots, \bm{x}_n\}$ are the model parameters to be estimated; 
\item\textit{factor nodes} $\mathcal{F}\triangleq\{l_1, \dots, l_u\}$  consist of the individual factors $l_k$; 
\item \textit{edge set} $\mathcal{E} \triangleq\{(\bm{x}_i, l_k) \in \mathcal{V} \times \mathcal{F}\,|\,\bm{x}_i\in\bm{x}_{S_k}\}$, i.e. variable $\bm{x}_i$ and factor $l_k$ are joined by an edge in $\mathcal{G}$ if and only if $\bm{x}_i$ is an argument of $l_k$.
\end{enumerate} 
See Fig. \ref{fig:certi-fgo}  for a representative example.

Factor graphs serve two primary functions. First, the edge set $\mathcal{E}$ in $\mathcal{G}$ directly encodes the sparsity structure of Prob.~\eqref{eq:fgo_opt_prob}. Second, factor graph models provide a convenient modular modeling language for constructing high dimensional optimization problems    by composing simple elementary constituent parts (i.e.\ individual variables and factors).  Consequently, many current state-of-the-art software libraries for state estimation in robotics and computer vision employ factor graph-based abstractions for instantiating and solving \gls{MLE} problems of the form \eqref{eq:fgo_opt_prob} \cite{agarwal2012ceres, gtsam}.

Now let us additionally suppose that the generic maximum likelihood estimation \eqref{eq:fgo_opt_prob} takes the form of the \gls{QCQP}~\eqref{eq:qcqp}.  Observe that the sparsity pattern captured in \eqref{eq:fgo_opt_prob} places algebraic restrictions on the \gls{QCQP} data matrices: $S_k$ captures the \textit{block sparsity pattern} of the $k$-th objective matrix $\bQ_k$, while the Cartesian product structure of the full domain implies constraint matrices $\bA_i$ are \emph{block-diagonal} with one nonzero block, i.e., they are \textit{block separable}.
\CertiFGO~\cite{certi_fgo} exploits this correspondence between sparse factor graphs and \gls{QCQP}s with block-structure by \textit{lifting} each {variable} to a higher-dimensional domain corresponding to the \gls{BM}-factored relaxation. Crucially, this lifting preserves the sparsity structure of \eqref{eq:fgo_opt_prob}.
Assuming that the lifted variable domains are smooth manifolds, the resulting \textit{lifted factor graph} can thus be optimized on the product manifold using Riemannian optimization \cite{boumal2023introduction}, where the Riemannian Staircase \cite{Boumal2015RS,papalia2024overview} (Algorithm~\ref{alg: certifiable_factor_graph}) automatically manages rank increases and solution verification. Thus, \CertiFGO provides certifiable global optimality and competitive scalability without requiring problem-specific implementations.
\begin{remark}\label{rmk:need_qcqp}
    The use of this method \textit{presupposes} that the estimation problem \eqref{eq:fgo_opt_prob} is a \gls{QCQP}. This is the case for many robotics and computer vision estimation problems \cite{SLAMHandbookTheoryChp}\cite{yang2020one}.
\end{remark}

In the following, we show how factor graph structure is preserved through the problem transformations (QCQP\(\rightarrow\)SDP\(\rightarrow\)BM) described in Secs.~\ref{subsec:shors} and \ref{subsec:low_rank_staircase}. Likewise, the verification and saddle-escape procedures \ref{subsec:low_rank_staircase} can be efficiently performed block wise by exploiting the same factor graph structure.

\subsubsection{QCQP and Shor's relaxation over Factor Graphs}
We begin by explicitly writing the factor graph \gls{MLE} problem~\eqref{eq:fgo_opt_prob} as a \gls{QCQP}~\eqref{eq:qcqp}, revealing the block separable structure of the objective and constraints induced by the factor graph formulation. The decision variable \(\bX\in\mathbb{R}^{n\times d}\) can be partitioned into \(K\) \textit{block-row variables} \(\bX_i\in\mathbb{R}^{d_i\times d}\) corresponding to individual factor graph variables $\bm{x}_i$ in~\eqref{eq:fgo_opt_prob}, such that  \(\bX=\begin{bmatrix}\bX_1&\bX_2&\cdots&\bX_K\end{bmatrix}^\top\) with \(n=\sum_{i=1}^K d_i\).

The decomposition \eqref{eq:fgo_opt_prob} places strong restrictions on the data matrices parameterizing the corresponding \gls{QCQP}.  First, recall that each factor $l_k(\cdot)$ only depends upon the subset of variables indexed by $S_k$.  This fact constrains the sparsity patterns of the data matrices $\bQ_k \in \mathbb{S}^n$ parameterizing the quadratic summands $l_k(\cdot)$ in \eqref{eq:fgo_opt_prob}: note that we must have $(\bQ_k)_{i,j} = 0$ if $(i,j) \notin S_k \times S_k$.  Second, the feasible set of \eqref{eq:fgo_opt_prob} is a \emph{Cartesian product} of the individual variable domains $\mathcal{M} = \mathcal{M}_1 \times \dots \times \mathcal{M}_K$, i.e., each variable can be varied independently of all others.  In order to be consistent with this product structure, it follows that each of the quadratic constraints appearing in \eqref{eq:qcqp} can involve only \emph{one} of the variables $\bX_i$.  Consequently, we may partition the index set $[m]$ of the constraints into subsets, where $L_i \subseteq [m]$ contains the indices of those constraints associated with variable $\bX_i$.  This implies that if $\ell \in L_i$, then $A_\ell$ is block diagonal, with a single nonzero block in the $(i,i)$-th position.  Thus, the factor graph decomposition \eqref{eq:fgo_opt_prob} implies the following block decomposition for the corresponding \gls{QCQP}:

\begin{equation}
\begin{aligned}
&\min_{\bX\in\mathbb{R}^{n\times d}} 
&&\sum_{k=1}^u \, \overbrace{\sum_{(i,j) \in S_k \times S_k}^K \langle (\bQ_k)_{i,j}, \bX_i\bX_j^\top\rangle}^{\ell_k(\bX)}
\label{equation: factor_graph_objective_as_QCQP}
\\
&\hspace{1em}\text{s.t.}
&&\langle(\bA_{\ell})_{i,i},\bX_i\,\bX_i^{\top}\rangle=b_{\ell},\;\forall \ell \in L_i,\ i\in[K],
\end{aligned}
\end{equation}

\subsubsection{Burer-Monteiro Factorization over Factor Graphs}
For scalability, we apply the low-rank \gls{BM} factorization $\bZ = \bY\bY^\top$ to the \gls{SDP} obtained with Shor's relaxation, partitioning $\bY \in \mathbb{R}^{n \times p}$ into $K$ variable blocks $\bY_i\in\mathbb{R}^{d_i\times p}$ to arrive at
\begin{equation}\label{equation:bm_fgo}
\begin{aligned}
\min_{\bY\in\mathbb{R}^{n\times p}} 
&\sum_{k=1}^u \, \overbrace{\sum_{(i,j) \in S_k \times S_k}^K \langle (\bQ_k)_{i,j}, \bY_i\bY_j^\top\rangle}^{\ell_k(\bY)}
\\
\text{s.t} \quad 
& \langle(\bA_{\ell})_{i,i}, \bY_i\bY_i^\top\rangle = b_\ell, \forall \ell \in L_i, i\in[K].
\end{aligned}
\end{equation}
%
Note that the sparsity and block-separability structure of the \gls{BM} factorization \eqref{equation:bm_fgo} \emph{exactly matches} the one assumed in the factor graph decomposition \eqref{eq:fgo_opt_prob} for the initial QCQP \eqref{equation: factor_graph_objective_as_QCQP}.

Rather than use traditional non-linear programming solvers to solve the constrained nonconvex problem~\eqref{equation:bm_fgo}, we leverage \emph{intrinsic optimization}—reformulating it as unconstrained optimization on manifolds—for substantial efficiency gains. The constraint set of Prob. ~\eqref{equation:bm_fgo} inherits the same block-diagonal and block-separability properties as that of the original QCQP~\eqref{equation: factor_graph_objective_as_QCQP}. It follows that the feasible set for our desired intrinsic reformulation must also be a Cartesian product, and that the individual factors comprising this product are determined by:

\begin{equation}
\begin{aligned} \label{eq:submanifold}
\mathcal{M}^{(p)}_{i}
:= &\bigl\{\bY_i\in\mathbb{R}^{d_i\times p} : \langle(\bA_{\ell})_{i, i}, \bY_i \bY_i^{\top}\rangle = b_\ell\bigr\},\\&\forall \ell \in L_i, i \in [K],
\end{aligned}
\end{equation}
with the overall feasible set $\mathcal{M}^{(p)}:= \mathcal{M}_1^{(p)} \times \cdots \times \mathcal{M}_K^{(p)}$. The lifted objective~\eqref{equation:bm_fgo} inherits the same sparse dependency structure from the original data matrices $\bQ_k$~\eqref{equation: factor_graph_objective_as_QCQP}, ensuring the reformulated problem maintains a factor graph representation with variables and factors in \textit{one-to-one} correspondence with the original \gls{MLE}~\eqref{eq:fgo_opt_prob}. This leads to the intrinsic formulation:

\begin{equation}\label{equation: intrinsic_optimization_formulation}
\min_{\bY\in\mathcal{M}^{(p)}}\;
\bigl\langle \bQ,\,\bY\,\bY^{\!\top}\bigr\rangle.
\end{equation}
In the (common) event that the constraints defining $\mathcal{M}^{(p)}$ in \eqref{eq:submanifold} determine a smooth manifold, the optimization \eqref{equation:  intrinsic_optimization_formulation} becomes a smooth unconstrained optimization over a product of smooth manifolds, enabling the application of Riemannian optimization methods from factor graph optimization libraries.

\subsubsection{Optimality Verification and Saddle-Escape in the Factor Graph Setting}
The Lagrange multipliers used in the optimality certificate of~\eqref{equation: definition_of_certificate} can be obtained by solving a least-squares problem.
Because each constraint matrix $\bA_\ell$ is block-diagonal with a single nonzero block $(\bA_{\ell})_{i, i}$ acting on variable $\bY_i$ for all $\ell\in L_i$, the operator $\mathcal{A}^\ast(\bm{\lambda})$ is also block-diagonal with blocks aligned to the factor graph variables. 
Consequently, the linear system characterizing Lagrange multipliers for the \gls{BM} factorization~\eqref{eq:burer_monteiro} actually decomposes into $K$ \textit{independent} linear systems involving the K blocks on the diagonal of $\mathcal{A}^\ast(\blambda)$.  It follows that we can solve for the Lagrange multipliers $\lambda_i$ (and hence the nonzero blocks of the certificate matrix $\bS$) by performing independent operations on the individual manifolds $\mathcal{M}_i^{(p)}$.
This blockwise structure enables efficient optimality verification and the construction of negative-curvature directions for saddle escape that directly reflect the sparsity of the underlying factor graph.

Such efficient optimality verification and saddle escape are carried out using the Riemannian Staircase algorithm~\cite{Boumal2015RS,papalia2024overview} adapted to the factor graph setting, described in Algorithm~\ref{alg: certifiable_factor_graph}. At each rank $p$, local Riemannian optimization of the \gls{BM} problem is performed over the product manifold 
$\mathcal{M}^{(p)}$, 
the certificate matrix $\bS$ is computed from the blockwise multipliers, and global optimality is either certified or the rank is increased along a negative-curvature direction to continue the search.

In practice, this block-separable formulation makes the Riemannian optimization problem almost automatic: the product manifold $\mathcal{M}^{(p)}$ can be assembled directly from the individual submanifolds $\mathcal{M}^{(p)}_i$ defined in~\eqref{eq:submanifold}, and the GNC scheme already implemented in GTSAM can be applied to the lifted factors without the need for problem-specific code. See~\cite{certi_fgo} for details on constructing these lifted factors.
%
As a result, practitioners can construct a robust solver that certifies global optimality of the inner WLS subproblems simply by lifting the original factor graph, rather than hand-deriving problem-specific data matrices or manifold models.

\begin{algorithm}[h]
\DontPrintSemicolon
\SetAlgoLined
\LinesNumbered
\caption{\CertiFGO, Certifiable Estimation in Factor Graphs~\cite{certi_fgo}}
\label{alg: certifiable_factor_graph}
\KwIn{Initial values $\bY = \{\bY_i\}_{i=1}^n$, factor graph $\mathcal{G}$, initial rank $p$.}
\KwOut{A feasible estimate $\hat{\bX}$ and lower bound $f^{\ast}_{\mathrm{SDP}}$ on its optimal value.}
\SetFuncSty{textnormal}
\SetKwFunction{FMain}{\textsc{CertifiableFGO}}
\SetKwProg{Fn}{function}{:}{end}

\Fn{\FMain{$\bY$, $\mathcal{G}$, $p$}}{%
    \While{true}{
        \Comment{Lift variables to rank-$p$}
        $\bY_p \gets \textsc{Lift}(\bY)$\;
        \Comment{Construct lifted factors for rank-$p$ lift}
        $\mathcal{G}_p \gets \textsc{ConstructLiftedFactors}(\mathcal{G}, p)$\;
        $(\bY^{\ast}_p) \gets \textsc{LocalOptimization}(\mathcal{G}_p, \bY_p)$\;
        \Comment{Construct certificate and verify by computing minimum eigenvalue}
        $(\lambda_{\min},v_{\min}) \gets \textsc{Verification}(\mathcal{G}_p, \bY_p)$\;
        \uIf{$\lambda_{\min} > 0$}{
            \Return{$\bigl\{\hat{\bY},\,f_{\!\mathrm{SDP}}^{*}\bigr\}$}\;
        }
        \Else{
            \Comment{Increase rank, use current solution as initialization in next iterate.}
            $p \gets p + 1$\;
            $\bY_{p} \gets \textsc{SaddleEscape}(\bY^{\ast}_p, v_{\min})$\;
        }
    } 
    $\hat{\bX} \gets \textsc{RoundSolution}(\bY_p)$\;
    \Return{$\bigl\{\hat{\bX},\,f_{\!\mathrm{SDP}}^{*}\bigr\}$}\;
} 
\end{algorithm}

\subsection{Robustifying Certifiable FGO}\label{sec:iii:implementation_details} 
In Sec.~\ref{sec:br}, we showed that the \gls{BR} duality reformulates M-estimation as iterative variable and weight updates (Algorithm~\ref{alg:algorithm_gnc}). The variable update is a \gls{WLS} problem—a type of \gls{QCQP}—which \CertiFGO can solve (see Remark \ref{rmk:need_qcqp}). Therefore, \CertiFGO naturally fits as the inner solver for adaptive reweighting schemes.
 
We emphasize how easy it is to \textit{implement} \CertiFGO in practice within existing factor graph-based adaptive reweighting robust estimation frameworks. Unlike purpose-built methods like SE-Sync \cite{rosen2019se} —which require hand-crafting the objective and constraint matrices for Problem~\eqref{eq:burer_monteiro}, manually specifying the manifold structure, and implementing custom solver subroutines (ex. gradient and Hessian computations)—\CertiFGO operates directly on factors. This seamless integration with GTSAM's mature solver library and \gls{GNC} implementation, provides robustified certifiable estimation for factor graphs  without manual problem reformulation and minimal implementation efforts. In this paper we present one realization of this pipeline which we refer to as \CertiGNC, consisting of Algorithm~\ref{alg:algorithm_gnc} with the inner \gls{WLS} variable update solved with Algorithm~\ref{alg: certifiable_factor_graph}.

At each rank-$p$ lift of \CertiFGO Algorithm~\ref{alg: certifiable_factor_graph}, optimization on the lifted manifold $\mathcal{M}^{(p)}$ as per~\eqref{equation: intrinsic_optimization_formulation} is performed using GTSAM's native Riemannian \gls{LM} solver, which handles the manifold geometry through tangent space computations and retractions without requiring any custom implementation. In our experiments, \verb+GNC-Local+—our non-certifiable \gls{GNC} baseline—uses this same \gls{LM} solver for its inner solves, ensuring observed performance differences stem from our use of certifiable inner \gls{WLS} solves, rather than from details of the underlying factor graph optimizers.

In practice, the inner loop of Algorithm~\ref{alg:algorithm_gnc} (lines 3-8) is truncated to a single iteration. When homotopy parameter updates are sufficiently small, the minimizer for the next step lies close to the current solution $\bm{x}^*$, allowing this warm-started single iteration to adequately approximate the solution. For \verb+GNC-Local+, this means performing one local solve per $\mu$ value. For \CertiGNC, each inner solve achieves global optimality, inherently eliminating any need for iterative refinement.

\textbf{Algorithm Termination.}
To ensure fair comparison, \CertiGNC and \verb+GNC-Local+ share the same \gls{GNC} \textit{outer loop} termination criteria. The loop terminates when any of the following occur between iterations: \textit{(i)} weights $w_i$ converge, i.e. $\max_{1\le i\le n}\bigl|\,w_i - \operatorname{round}(w_i)\,\bigr| \le \varepsilon$, \textit{(ii)} cost converges, i.e. $\Delta \langle \bQ, \bY\bY^{\top}\rangle \leq c_{tol} = 10^{-6}$, \textit{(iii)} the max number of GNC iterations reached.
With small $\mu$ steps, a single warm-started WLS per stage often suffices and the inner loop may be omitted.
Similarly, the exit criterion for the \textit{inner loop} \gls{WLS} solves are (\CertiFGO or LM solve of \verb+GNC-Local+ ): \textit{(i)} cost converges, i.e. $\Delta \langle \bQ, \bY\bY^{\top}\rangle \leq  c_{tol} = 10^{-5}$ 
, \textit{(ii)}  $p_{max} = 30$, maximum iterations exceeded, \textit{(iii)}  $S + \eta \mathbf{I} \succcurlyeq 0$, i.e. solution certification achieved. Note that criterion \textit{(iii)} only applies to
the Riemannian Staircase of \CertiFGO in Algorithm~\ref{alg: certifiable_factor_graph}.

\begin{remark}\label{rmk:strictness}
\CertiFGO performance depends on parameters $\eta$ and $p_{max}$, which can be tuned to balance certificate accuracy and runtime. While smaller $\eta$ values give stricter certificates, in practice using larger $\eta$ values for approximate certificates can, in some instances, still achieve near-optimal solutions. While the maximum rank $p_{max}$ affects solver speed, certificates are usually obtained at lower ranks than the specified maximum.
\end{remark}

\section{\gls{FGO} Applications and Experiments}

\begin{figure*}[t]
    \centering
    \includegraphics[width=1\textwidth]{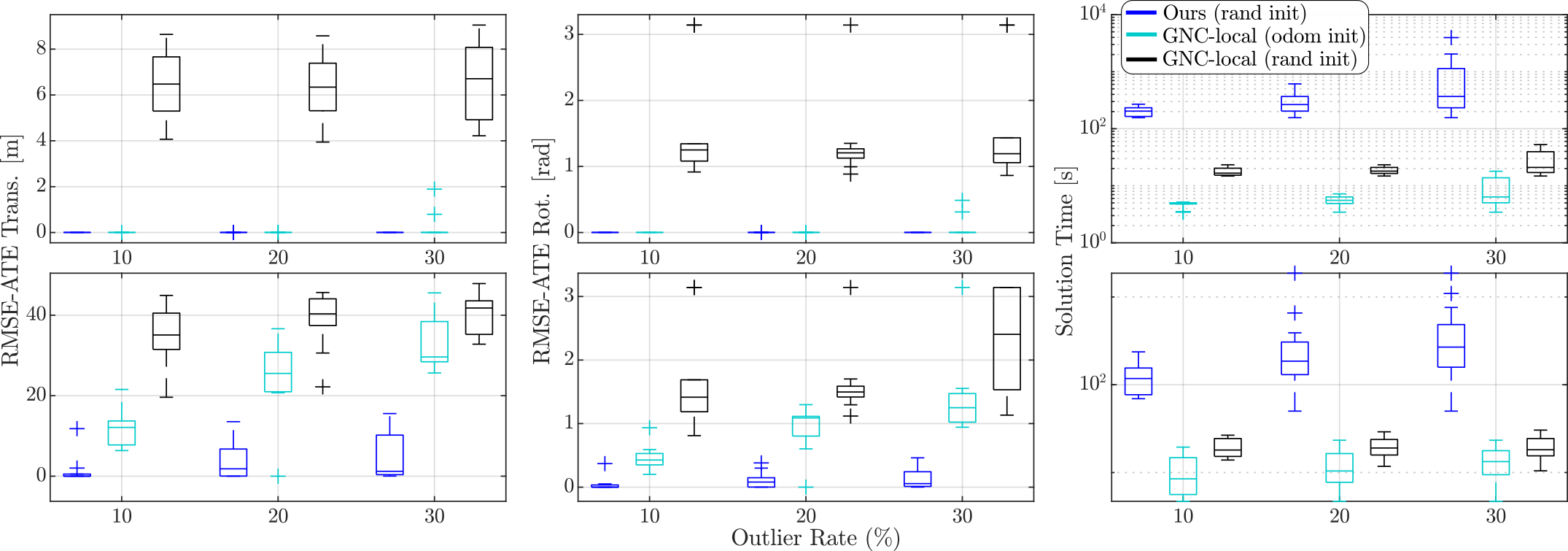}
    \caption{Performance of our \CertiGNC framework with random initialization compared with the non-certifiable baseline \GNCLocal given a good initialization from odometry and a random one. 
    Columns (left to right): RMSE-ATE (translation), RMSE-ATE (rotation), and solution time. 
    Rows (top to bottom): a) PGO problems from Intel dataset and b) landmark \gls{SLAM} from Trees dataset \cite{kaess2012isam2}.
    }
    \label{fig:pgo_results}
\end{figure*}

We benchmark \CertiGNC, our iterative framework with globally optimal subproblem solves that requires no good initialization (see Fig.~\ref{fig:intro}), against a purely local one \GNCLocal under two initializations: random sampling on the feasible set (rand init) and a favorable (outlier-free) odometry-based initialization (odom init).
We evaluate two representative factor graph problems: \gls{PGO} and landmark-based \gls{SLAM}.

\subsection{Experimental Setup}\label{sec:iii:experimental_setup}

\textbf{Problem Definition:} The standard \gls{PGO} problem estimates poses \(\{(\mathbf{R}_{i},\mathbf{t}_{i})\}_{i\in[K]}\), with \(\mathbf{R}_{i}\in\mathrm{SO}(d)\) and \(\mathbf{t}_{i}\in\mathbb{R}^{d}\), from noisy relative measurements \(\{(\tilde{\mathbf{R}}_{ij},\tilde{\mathbf{t}}_{ij})\}_{(i,j)\in\mathcal{E}}\) by minimizing \(\sum_{(i,j)\in\mathcal{E}}\kappa_{ij}\|\mathbf{R}_{j}-\mathbf{R}_{i}\tilde{\mathbf{R}}_{ij}\|_{F}^{2}+\tau_{ij}\|\mathbf{t}_{j}-\mathbf{t}_{i}-\mathbf{R}_{i}\tilde{\mathbf{t}}_{ij}\|_{2}^{2}\), where \(\kappa_{ij}\) and \(\tau_{ij}\) are the measurement precisions. Landmark-based \gls{SLAM} extends \gls{PGO} by estimating landmark positions \(\{\mathbf{l}_{k}\}_{k\in[L]}\) from noisy pose–landmark measurements \(\{\tilde{\mathbf{l}}_{ik}\}_{(i,k)\in\mathcal{E}_{\text{lm}}}\), with \(\mathbf{l}_{k}\in\mathbb{R}^{d}\), and adds the residuals \(\sum_{(i,k)\in\mathcal{E}_{\text{lm}}}\tau_{ik}\|\mathbf{l}_{k}-\mathbf{t}_{i}-\mathbf{R}_{i}\tilde{\mathbf{l}}_{ik}\|_{2}^{2}\), where \(\tau_{ik}\) is the measurement precision.

\textbf{Metrics:} We report runtime (in seconds) and estimation accuracy, the latter measured by the \gls{RMSE-ATE}~\cite{zhang2018tutorial}.

\textbf{Implementation Details:} 
For outlier generation, we corrupt only loop-closure (PGO) or pose–landmark (landmark SLAM) edges at rates (10\%, 20\%, 30\%) by introducing randomly generated edges, mimicking data association failures; odometry measurements are left unperturbed and thus remain inliers.
Each problem is evaluated over 10 Monte Carlo trials with independently generated outlier realizations.
\CertiGNC and \GNCLocal use the same GNC parameters as introduced in the previous section.
The GTSAM \gls{LM} solver is used as the local solver for both \CertiGNC and \GNCLocal, with  \texttt{relative\_error} \(10^{-8}\), \texttt{absolute\_error} \(10^{-8}\), and \texttt{max\_iteration} \(100\).
Note that the parameters can be tuned to achieve improved performance across different randomly generated outlier patterns.
All experiments use the certifiable \gls{FGO} implementation of~\cite{certi_fgo} within the \gls{GNC} scheme from GTSAM~\cite{gtsam, yang2020graduated}, and were run on a laptop equipped with an Intel Core i7-11800H CPU and 32\,GB RAM under Ubuntu~22.04. For \textsc{Trees} (landmark SLAM), we evaluate a reduced variant, extracting the first 1600 poses and the corresponding landmark observations for those poses.

\subsection{Accuracy and Runtime Results}\label{sec:iii:results}

We assess whether \CertiGNC provides practical benefits over local methods for state estimation problems of varying complexity in the presence of outliers. The results in Fig.~\ref{fig:pgo_results} reveal two key findings.

In terms of translation and rotation error, and for the easier estimation problem of PGO, \GNCLocal with good odometry initialization and \CertiGNC provide similar results, though \GNCLocal exhibits higher variability, indicating less reliable performance. For the more challenging landmark SLAM problem, \CertiGNC consistently outperforms \GNCLocal across all conditions, showing that global optimization guarantees become essential as problem complexity increases due to higher dimensionality and more opportunities for local methods to fail.

While \CertiGNC exhibits longer convergence times than local alternatives, this is expected given the additional computational complexity of solving the underlying global optimization problem and the certification framework's computational overhead~\cite{rosen2022accelerating} (see Remark~\ref{rmk:strictness}). This is justified by the improved solution reliability.

\subsection{Tightness Properties Under Outlier Contamination}\label{sec:iii:case_study}

For certain estimation problems \cite{rosen2019se, fan2020cpl}, SDP relaxations are known to remain tight up to a problem-dependent noise level, in which case they yield exact, low-rank solutions (see Remarks~\ref{rem:solution-recovery},~\ref{rem:low-rank-justification}).  
These properties enable the use of \gls{BM} factorization to achieve computational tractability. However, in robust estimation the presence of outliers violates this bounded-noise assumption. Thus, we cannot generically expect \gls{FGO} problems with outliers to exhibit the low-rank structure and exactness that \CertiFGO is designed to exploit.
On the other hand, if the iterative reweighting procedure used in \gls{GNC} succeeds in effectively separating inliers from outliers, we should hope that the \emph{reweighted} inner least-squares estimation problems produced at the end of the homotopy will recover these tightness properties.

\begin{figure}\label{fig:}
\centering
        \includegraphics[width=1\columnwidth]{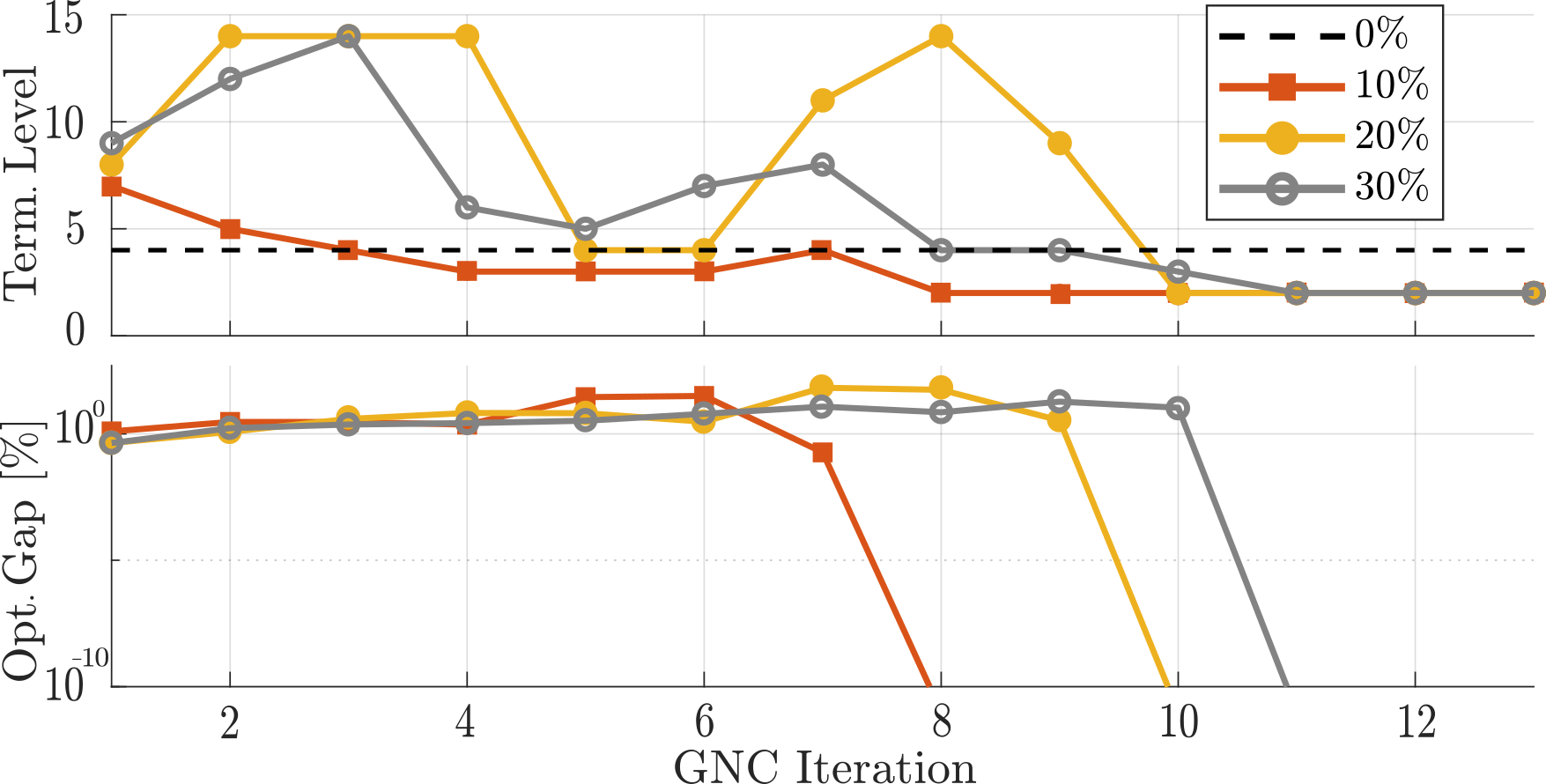}
        \caption{A single Monte Carlo trial on the Intel \gls{PGO} dataset with outlier rates of $0$\%, $10$\%, $20$\%, and $30$\%, showing (top) the Riemannian-staircase termination rank (a proxy for computational effort) and (bottom) the stagewise optimality gap of the \CertiFGO solve at each \gls{GNC} iteration. In the $0$\% outlier case, \gls{GNC} weights converge in the first iteration and the optimality gap is $10^{-12}$ as per Remark \ref{rem:solution-recovery}, the terminal level is visualized with a dashed line. All trials start at level~2 with random initialization; the trials with outlier terminated at \gls{GNC} iteration~13.} \label{fig:rank}
        \vspace{-0.5cm}
\end{figure}

For a set of representative Monte Carlo \gls{PGO} samples subject to varying outlier contamination, the termination level and relative suboptimality gap \(\frac{f_{\mathrm{QCQP}}-f_{\mathrm{SDP}}}{f_{\mathrm{SDP}}}\) of the \CertiFGO solution, embedded within each \gls{GNC} iteration, are plotted in Fig. \ref{fig:rank}. Across all iterates the termination level (Fig. \ref{fig:rank}, top) remains relatively small and converges to a low value, demonstrating that the requisite low-rank structure is indeed present and that \CertiFGO successfully exploits this. Since the termination level directly corresponds to computational effort, the maximum level of $p=15$ indicates that \gls{BM} factorization provides scalability even when used as an inner solver in outlier-contaminated settings. 
Similarly, the relative suboptimality  gap approaching $0$ (Fig. \ref{fig:rank}, bottom) indicates that our non-convex \gls{SDP} relaxations become asymptotically tight at the termination of GNC (i.e., after finding suitable assignments to the weights $w_i$), confirming the recovery of the low-rank structure and exactness that our method is designed to exploit.

\section{Conclusion}

In this paper, we show how to implement \textit{robustified} certifiable estimators using only fast \emph{local} optimization over smooth manifolds, thus enabling practitioners to easily design and deploy these state-of-the-art methods using standard factor graph-based software libraries and workflows.
We embed the recently proposed non-robust certifiable factor graph estimation framework in~\cite{certi_fgo} as the inner solver within an robust adaptive reweighting scheme, GNC, providing stage-wise global optimality certificates for the inner subproblems. 

In terms of estimation accuracy, our framework exceeds the performance of M-estimation strategies based on local search. While our method is more expensive to run than local baselines, this is expected because we are performing \textit{global} rather than \textit{local} optimization for the inner \gls{WLS} solves. This trade-off is worthwhile for the improved reliability our approach affords. 
Finally, by requiring no problem-specific relaxations and leveraging standard facto graph software (e.g.\ GTSAM), our approach facilitates broad deployment across diverse estimation tasks, thereby democratizing access to the powerful machinery of robust certifiable estimation.

\bibliographystyle{IEEEtran}
\bibliography{ref}

\end{document}